\def\year{2021}\relax
\documentclass[letterpaper]{article} % DO NOT CHANGE THIS
\usepackage{aaai21}  % DO NOT CHANGE THIS
\usepackage{times}  % DO NOT CHANGE THIS
\usepackage{helvet} % DO NOT CHANGE THIS
\usepackage{courier}  % DO NOT CHANGE THIS
\usepackage[hyphens]{url}  % DO NOT CHANGE THIS
\usepackage{graphicx} % DO NOT CHANGE THIS
\usepackage{amsmath}
\usepackage{amssymb}
\usepackage{threeparttable}
\usepackage{graphicx}  % DO NOT CHANGE THIS
\usepackage{natbib}  % DO NOT CHANGE THIS AND DO NOT ADD ANY OPTIONS TO IT
\usepackage{caption} % DO NOT CHANGE THIS AND DO NOT ADD ANY OPTIONS TO IT
\usepackage{color}
\usepackage{colortbl}
\usepackage{diagbox} 
\usepackage{multirow}
\usepackage{booktabs}
\usepackage{subfigure} 
\usepackage{url}
\usepackage[pagebackref=true,breaklinks=true,colorlinks,citecolor=blue,bookmarks=false]{hyperref} 

\title{Regional Attention with Architecture-Rebuilt 3D Network\\
for RGB-D  Gesture Recognition}
\author {
        Benjia Zhou$^{1}$,
        Yunan Li $^{3,4}$,\thanks{Benjia Zhou and Yunan Li contribute equally to this paper.}
        Jun Wan $^{2,5}$\thanks{Corresponding author;
        This work was done by Benjia Zhou who visited the lab of Center for Biometrics and Security Research, NLPR, CASIA.
        } \\
}
\affiliations {
    $^{1}$Macau University of Science and Technology, Macau SAR, China\\
$^{2}$ National Laboratory of Pattern Recognition, Institute of Automation, Chinese Academy of Sciences, Beijing, China\\
$^{3}$ School of Computer Science and Technology, Xidian Univeristy, China\\
$^{4}$ Xi'an Key Laboratory of Big Data and Intelligent Vision, China \\
$^{5}$ School of Artiﬁcial Intelligence, University of Chinese Academy of Sciences, Beijing, China \\

    {19098536ii20001@student.must.edu.mo, yunanli@xidian.edu.cn, jun.wan@ia.ac.cn}
    
}
\begin{document}
\def\etal{\emph{et al}.}
\def\eg{\emph{e}.\emph{g}.}
\def\ie{\emph{i}.\emph{e}.}
\maketitle
\begin{abstract}
Human gesture recognition has drawn much attention in the area of computer vision. However, the performance of gesture recognition is always influenced by some gesture-irrelevant factors like the background and the clothes of performers. Therefore, focusing on the regions of hand/arm is important to the gesture recognition. Meanwhile, a more adaptive architecture-searched network structure can also perform better than the block-fixed ones like ResNet since it increases the diversity of features in different stages of the network better.
In this paper, we propose a regional attention with architecture-rebuilt 3D network (RAAR3DNet) for gesture recognition.
% Noticing the features in shallow and deep layers of the network  show different relations with the recognition task, 
We replace the fixed Inception modules with the automatically rebuilt structure through the network via Neural Architecture Search (NAS), owing to the different shape and representation ability of features in the early, middle, and late stage of the network. 
%{\color{red}{Specifically, we make the network rebuild structure of modules automatically through the network with the technique of Neural Architecture Search (NAS),}}
%{\color{red}{which implicates the different relations.}}
It enables the network to capture different levels of feature representations at different layers more adaptively. Meanwhile, we also design a stackable regional attention module called Dynamic-Static Attention (DSA), which derives a Gaussian guidance heatmap and dynamic motion map to highlight the hand/arm regions and the motion information in the spatial and temporal domains, respectively. Extensive experiments on two recent large-scale RGB-D gesture datasets validate the effectiveness of the proposed method and show it outperforms state-of-the-art methods.
The codes of our method are available at:  https://github.com/zhoubenjia/RAAR3DNet.

\end{abstract}

\section{Introduction}

Gesture is produced as part of deliberate action and signs, involving the motion of the up body, especially the arms, hands, and fingers. Video-based classification makes an essential component in gesture recognition. It has been applied to many human-centred tasks, such as apparent personality analysis \cite{li2020cr}, sign language recognition \cite{cui2019deep} and human-computer interaction (HCI) \cite{wang2016largei}. The handcrafted features \cite{wan2013one, wan2015explore} are always used for gesture recognition in the early years. The powerful feature representation ability of deep learning also promotes the application of neural networks in the field of gesture recognition  \cite{li2016large,miao2017multimodal,simonyan2014two,narayana2018gesture}.

For most of the deep learning-based gesture recognition methods, some popular networks like ResNet \cite{he2016deep}, SENet \cite{hu2018squeeze} and Inflated 3D Network (I3D) \cite{carreira2017quo} are usually employed as the backbone for gesture recognition. %\cite{miao2017multimodal}.
Although these networks have achieved great success in many tasks, it still worth pointing that the same modules are shared from shallow to deep layers in these networks. Even the modules in networks like I3D that employ multi-branch structure to improve the width and diversity are fixed and all the same through the network. However, features in the early stage and late stage are quite different. Features in the early stage are low-level features, which show the visual texture in each frame, whereas the high-level features in the late stage are abstracted and more related to the class of gestures. Therefore, it is not suitable to use the same structure to learn different features, and then we need to make the network more adaptive and automatically determine what the shape is for different parts of it.
%{\color{red}{This phenomenon}} also occurs in the condition that the different branches in the inception-like modules\cite{inceptionv1,inceptionv2}. 

%In the past decades, many approaches are proposed, ranging from handcrafted feature-based \cite{konecny2014one,malgireddy2012temporal,malgireddy2013language,wan2015explore} to deep learning-based methods\textcolor[rgb]{1,0,0}{\cite{karpathy2014large,ji2017spatial,liu2017continuous,qiu2017learning}}. With the release of Kinect and Intel Resense, both RGB and depth images are available. It leads to some multi-modal datasets~\cite{wan2016chalearn,Shahroudy2016CVPR,molchanov2016online} release and pushes the state-of-the-art techniques~\cite{li2016large,miao2017multimodal,simonyan2014two,zhang2018attention,narayana2018gesture} for multi-modal gesture recognition.

\begin{figure*}[ht]
\centering
\includegraphics[width=0.95\linewidth]{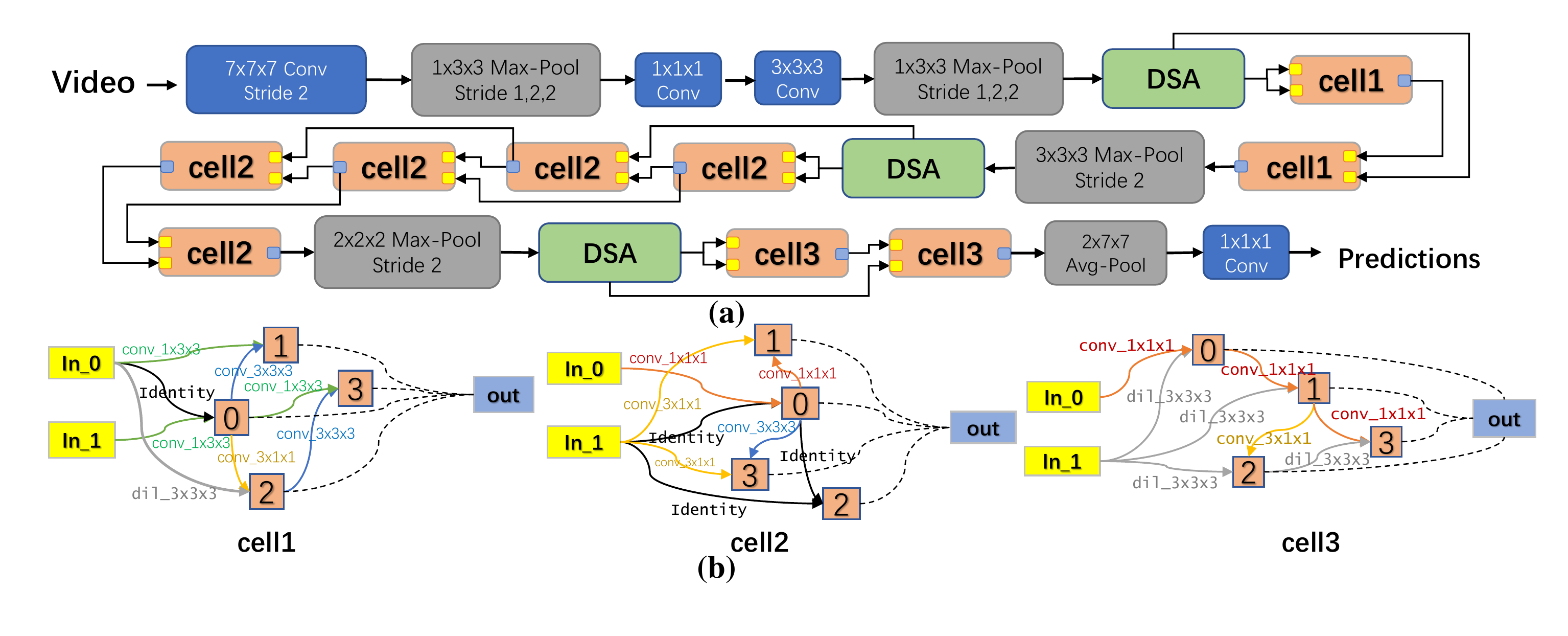}
\caption{The pipeline of the proposed method. (a) the structure of  the entire RAAR3DNet. (b) the inner structure of automatically rebuilt cell1, cell2, and cell3. 
Each cell is composed of two input nodes, four intermediate nodes and one output node. The output node is obtained by combing features from intermediate nodes with some reduction operation (\eg, concatenation), which are marked as the dashed lines in (b).
We take the I3D network as the backbone, and utilize NAS to automatically rebuild the structure of Inception Modules in it.  The reconstructed network shows different structure to fit multi-scale features. 
Cell1 and cell2, which are in the early and middle stage of the network, tend to employ convolution kernels with small receptive fields to capture the low-level texture features more easily, whereas cell3s at the end of the network perform dilated convolution operations to capture the more abstract and semantically high-level features.
%Cell1 and cell2, which are in the early and middle stage of the network, tend to employ convolution kernels with larger receptive fields to capture the low-level texture features more easily, whereas cell3 at the end of the network almost have no convolution operation, and in this way these modules can directly pass the abstract and semantically high-level features to the classifier.
}
%We found that the cell structures located in the shallow and middle layers seem to be more inclined to use convolution kernels with larger receptive fields, such as cell1 and cell2. This may be because neurons located in the shallow layer tend to be more sensitive to texture features that are easier to capture.  For neurons located in the deep layer, we hope that they can aggregate the features extracted by the shallow and middle layers, so that they can express more abstract and semantically rich information since these information can often determine the final classification performance. So as shown in cell3, we found that this layer of network has almost no convolution operation in the spatial dimension. }}}NASA3D
% \caption{The pipeline of our network. We take the I3D as the backbone network, and use NAS to reconstruct the internal Inception structure to stimulate it to achieve the best performance in specific tasks(\ie, gesture recognition tasks in this paper).}
\label{fig:pipeline}
\end{figure*}

Meanwhile, one of the most significant challenge hindering the improvement of recognition accuracy is the influence of gesture-irrelevant factors, such as backgrounds, different clothes of performers, and so on. The various textures and appearances could mislead the network to learn inconsequential or less important features.  For dynamic gesture recognition from a video sequence, we believe it is vital to focus on gesture movements, such as hands, arms or elbows of the performers. %Therefore, avoiding the misleading of gesture-irrelevant factors and focusing on gesture regions are required. 
%Another concern is how to make the network more easily trained and applied to different situations.
Many researchers notice that it is critical to make the network focusing on the gesture regions both spatially and temporally. Modules such as hand detector~\cite{wang2016largec,liu2017continuous}, and additional modalities of data like optical flow \cite{li2017largea} or saliency \cite{li2017largeb} are widely used via  combining with the raw RGB (and depth) data to design different algorithms. However, most of them require extra offline operations  (\eg, hand detection, optical flow calculation) in advance.  It would increase time complexity because of using hand detector network in the testing stage. Therefore, it may be more reasonable if the attention maps of gesture regions are learned along with the task of gesture recognition in the same network. 

%In addition, many discovering state-of-the-art neural network architectures are designed by experts through continuous efforts and exploration, such as SENet , ResNet \cite{he2016deep}  and I3D. Although these networks have achieved great success in many tasks, but for specific tasks, such as gesture recognition, these networks hardly capture tiny motion information, and it is hard to continuously improve these network architectures. Interestingly, in recent years, from the initial discrete architecture search method to the current continuous architecture search method, the Neural Architecture Search (NAS) algorithm has become more and more popular, and the accuracy and search efficiency have made a qualitative leap, which inspires us whether we can use NAS to search several local sub-network structures on these artificially designed network architectures to fit specific tasks, so that the performance of these classic networks on specific tasks can be maximized?

%Meanwhile, for multi-modal gesture recognition, the complementary feature learning can be benefited from different data modalities in different aspects. For example, with the depth data, it can be easy to distinguish foregrounds (\ie, face, hands and arms) from backgrounds while RGB data can provide more texture/color appearances. These features can benefit each other. Thus the results of different modalities of data can be fused in the network rather than trained separately and combined as late fusion.

Inspired by the above discussions, we propose a regional attention with architecture-rebuilt 3D network for dynamic gesture recognition based on RGB-D data, which is illustrated in Fig.\ref{fig:pipeline}. We take the I3D\footnote{We still utilize a two-stream configuration - with one I3D network trained on RGB inputs, and another on depth inputs.} network as the backbone and employ the theory of NAS to find the optimal combination of different operations in each module of the network.
To make the network focus on the gesture regions, we propose a regional attention module DSA, which includes a static attention sub-module (SAtt) and dynamic attention sub-module (DAtt).  % called ``dynamic-static Attention (DSA)'' module.
For static attention, we learn a heatmap of hands or body for each frame with the supervision of the Gaussian map of skeleton keypoints. It indicates where the hands/arms are and highlights these regions.
For dynamic attention, we present a fast approximate rank pooling algorithm to learn the accumulated dynamic images, which reduces the time complexity a lot when compared with the traditional rank pooling techniques \cite{bilen2016dynamic,bilen2017action} and thus can give a real-time dynamic image computation.
Then with the DSA structure applied, the network can pay attention to the gesture regions spatiotemporally.
%Meanwhile, in order to exploit the relationship among different modalities of data, we propose an adaptive cross-modal weighting (ACmW) module, which is employed in each residual block throughout the network to combine the information of multi-modal data. Unlike the one-off fusion strategies\cite{wang2016largei,li2016large}, inspired by message passing mechanism of FishNet \cite{sun2018fishnet}, we use the ACmW module to generate the ``correlation message'' of different modalities, and combine it with the original data stream for the successive feature learning. \textbf{This strategy can avoid losing details of single modality of data in the early stage.}  With the above two designs, the network can be end-to-end trainable and focus on gesture related features even with multi-modal inputs. 
Our contributions can be summarized as three-fold:

% \begin{figure*}[!tbp]
% \centering
% \includegraphics[width=1.0\linewidth]{img/pipeline_raw.jpg}
% \caption{The pipeline of our network. We take the Inception-V1 as the backbone network, and use NAS to reconstruct the internal Inception structure to stimulate it to achieve the best performance in specific tasks(\ie, gesture recognition tasks in this paper).}
% \label{fig:pipeline}
% \end{figure*}

% \begin{figure*}[!tbp]
% \centering
% \includegraphics[width=1.0\linewidth]{img/cells.jpg}
% \caption{The searched different cells from the  NAS.}
% \label{fig:cells}
% \end{figure*}

(1) We replace the structure-fixed modules in the general network with automatically reconstructed cells via NAS. The cells in the early, middle, and late stages of the network can have different structures and learn the low-level and high-level features more adaptively.
%We employ NAS to automatically find the optimal architecture of modules for different stages of the network. It distinguishes the structure of early stages from the later ones to better learn the low-level and high-level features.

(2) We propose a stackable attention structure, called DSA, to generate attention map in both spatial and temporal space. DSA consists of the SAtt and DAtt sub-modules. SAtt highlights the hands/arms features via an online learnable Gaussian skeleton heatmap while DAtt captures the gesture motions via the proposed fast approximate rank pooling algorithm with decreasing the time complexity to a large extent.

%(3) We propose an adaptive fusion module called Adaptive Cross-modal Weighting (ACmW). Unlike the previous offline multimodal fusion scheme \cite{miao2017multimodal,narayana2018gesture}, ACmW enables the network to end-to-end train different modalities of data together and leverages the strength of different modalities of data by generating the ``correlation message'' of them and combine it with each feature stream instead of simply late-fuse the score of different data.

(3) Extensive experiments that prove the integration of our designs can ultimately improve the performance of gesture recognition. Experiments demonstrate that our method can strike the balance between good performance and low computation burden, and outperform those top techniques on two large-scale gesture datasets.

\section{Related Work}

\subsection{Evolution of Approaches for Gesture Recognition}
The study on gesture taxonomies and representations has been continued for many years. Early methods are often based on handcrafted features \cite{klaser2008spatio,wan20143d}. 
Recently, the rapid progress of deep learning boosts many deep neural network-based feature extraction methods. 
The 2D convolutional neural network (CNN) and its derivations combining different branches for RGB and optical flow data \cite{simonyan2014two} are first used for gesture/action recognition tasks. Then some works~\cite{tran2015learning,li2017largea,carreira2017quo} use the 3D CNN for recognizing gestures, whereas some other methods~\cite{zhu2017multimodal,zhang2018attention} employ LSTM and its variants to model the temporal relationships for the gestures. There are also some methods \cite{wang2016largei,wang2018cooperative} exploited the features like dynamic images instead of the raw RGB-D data and used them as the inputs for gesture recognition.
Besides the RGB-D data, some other modalities of data like optical flow \cite{li2017largeb,miao2017multimodal} or saliency video \cite{li2017largea,duan2018unified} are also employed for improving the performance.

%The study on gesture taxonomies and representations has been continued for many years. For video-based gesture recognition, tracking the motion of hands and arms is a difficult task owing to the large of degree (DOF). Many handcrafted features \cite{klaser2008spatio,sanin2013spatio,wan20143d,wan2015explore} and deep neural network-based feature extraction methods \cite{tran2015learning,carreira2017quo,qiu2017learning} are proposed to try to solve this problem.
%In recent years, owing to the release of RGB-D sensors, the simultaneously captured RGB and depth data are available. These two kinds of data can be complementary to each other, and correspondingly help recognize gestures more accurately.
%Some works~\cite{tran2015learning,miao2017multimodal,li2017largea} use the 3D CNN for recognizing gestures, whereas some other methods~\cite{zhu2017multimodal,zhang2018attention} employ LSTM and its variants to model the temporal relationships for the gestures. There are also some methods \cite{wang2016largei,wang2016largec,wang2017scene,wang2018cooperative} exploited the features like dynamic images instead of the raw RGB-D data and used them as the inputs for gesture recognition.
%Besides the RGB-D data, some other modalities of data like optical flow \cite{li2017largeb,miao2017multimodal} or saliency video \cite{li2017largea,wang2017largei,duan2018unified} are also employed for improving the performance.

\subsection{Neural Architecture Search in Action Recognition}
Our work is driven by Neural Architecture Search (NAS)~\cite{liu2018darts,xu2019pc}. The development of NAS can be summarized in three branches: 1) based on reinforcement learning\cite{zoph2018learning}. 2) based on evolution \cite{real2019regularized,real2017large} and 3) based on gradient \cite{liu2018darts,xu2019pc}. Thanks to the recent high-efficiency and high-precision search methods of NAS, many NAS-based single-modal and multi-modal methods have been applied in action and gesture recognition tasks~\cite{zhang2020sar, yu2020searching}. Peng \etal\cite{peng2019video}  first attempted to automatically design a neural network for video action recognition tasks through NAS. At the same time, in order to reduce search costs, they introduced a temporal segmentation method that reduced the computational cost without losing global video information. Qiu \etal\cite{qiu2017learning} proposed Scheduled Differentiable Architecture Search (DAS), which can efficiently and automatically explore the network structure through gradient descent in images and videos. P{\'e}rez-R{\'u}a \etal\cite{perez2019mfas} proposed a general multi-modal neural architecture search method (MFAS) , which solved the problem of finding a good architecture for multi-modal classification problems. Wang \etal \cite{wang2020attentionnas} propose a novel search space for spatiotemporal attention cells (AttentionNAS) for video classification tasks. Different from the work of others, we utilize NAS to make the network automatically rebuild the structure of modules in the I3D to capture different levels of feature representations at different layers more adaptively.

\begin{figure}[ht]
\centering
\includegraphics[width=0.8\linewidth]{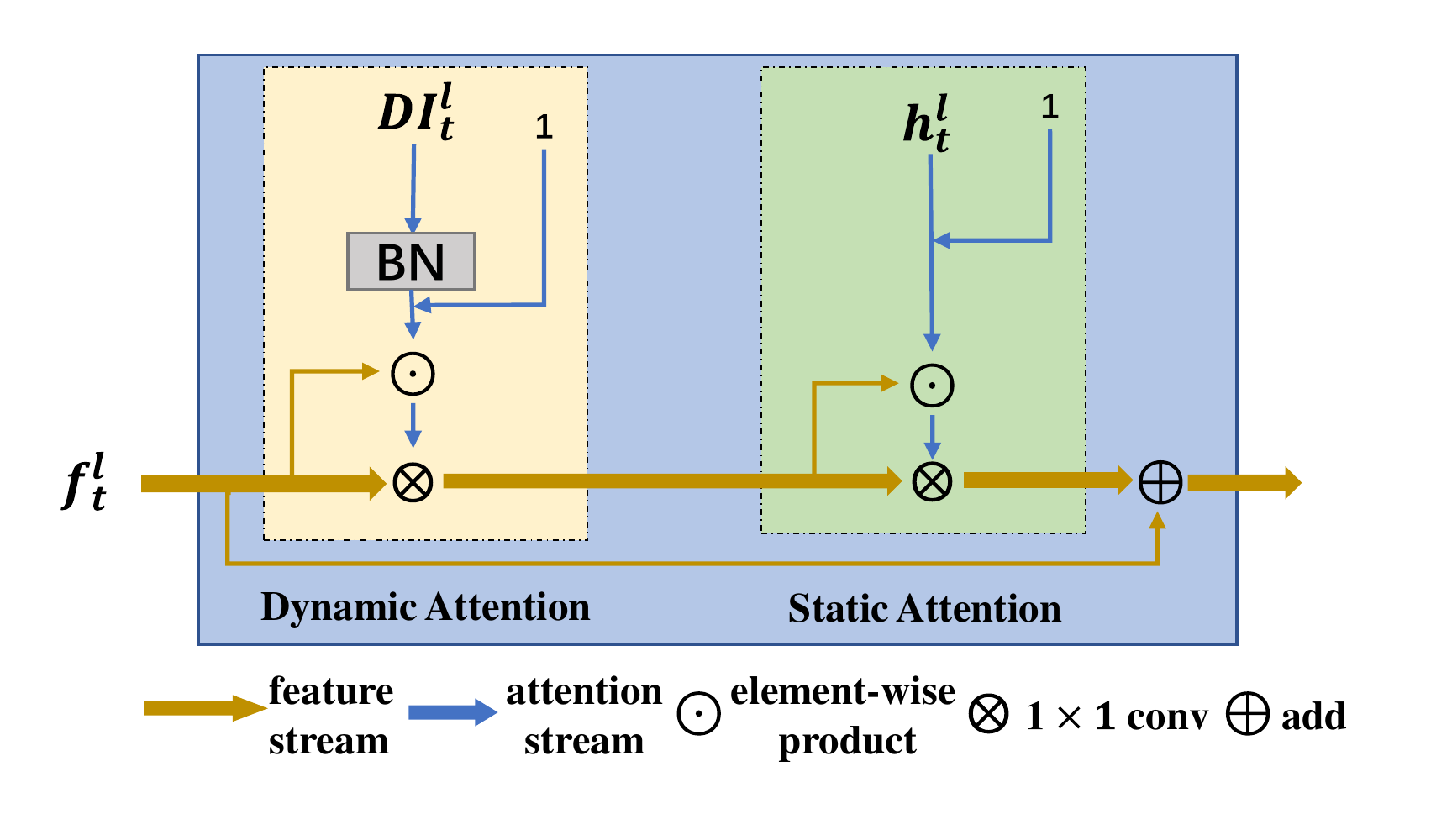}
\caption{The detail of DSA module. It has two sub-modules of dynamic attention (DAtt) and static attention (SAtt), which are sequentially combined together.}
\label{fig:hsatt}
\end{figure}

\subsection{Attention Mechanism in Gesture Recognition}
Attention mechanism has been wildly used in both low-level and high-level tasks like pose estimation \cite{chu2017multi}, object detection \cite{li2019zoom} and image restoration \cite{li2019lap}.
As the interference of backgrounds, the clothes of performers and the diversity of presentation for the same gesture are still the barrier for improving the recognition accuracy, many researchers also employ the attention mechanism to guide the network to focus on the gesture itself through the video. Some methods concentrate on the regions of gesture in each frame. Liu \etal \cite{liu2017continuous} leverage the faster R-CNN \cite{ren2015faster} as the hand detector to highlight the corresponding regions.  Lin \etal~\cite{lin2018large} utilize both detected hands and skeleton information to further focus on the gesture. The work~\cite{narayana2018gesture} uses a focus of attention network (FOANet) to extract global raw data, local left and right hand regions via different networks. The other ones mainly concern about the motion information among frames. In \cite{li2017largeb,miao2017multimodal}, the additional modality of optical flow data is used to capture the movements in videos.  Zhang \etal \cite{zhang2018attention} explore attention modules used in different gates of the LSTM when incorporating it and the 3D CNNs \cite{tran2015learning}. 
Although with these techniques the network can focus on the gestures, extra modalities of data served as an input of the network or offline training process for models like hand detector are always inevitable. On the contrary, the attention module of DSA in our method does not require any other data except for the RGB-D ones from the original dataset. The module can also be learned end-to-end within the recognition network, and it makes the network easier to apply to different situations with less time complexity. Meanwhile, since the DSA highlights both body parts in each frame and the movements through the adjacent frames, it facilitates the advantage of the temporal and spatial attention concurrently.

\section{Proposed Method}
\subsection{Overview of the Network}
The pipeline of RAAR3DNet is shown in Fig.\ref{fig:pipeline}. The network takes the I3D network as the backbone, and leverages NAS to automatically find the optimal structure for early, middle and late stage of the network to replace the original shape-fixed Inception Modules. Meanwhile, to better focus on the location and movement information of gesture-related parts like hands and arms, as show in Fig.\ref{fig:hsatt}, a stackable regional attention module DSA is embedded in the network. 
%\ie, we searched for three types of cell structures to replace inception module in the I3D network. %The DSA module helps to indicate where the hands/arms are and show the motion trajectory among frames.

% For searching the structure inside different modules of the network, we {\color{red}{construct a learnable weight matrix for each type of cell, where each row indicates the weight of each operator between pair of nodes. We continuously update this weight matrix through cross-entropy loss, and finally take the operation with the largest weight as the final operation. More details will be described in the second subsection below.}}

% In the DSA module, two kinds of attention are learned by two sub-modules of dynamic attention and static attention, respectively. In the dynamic attention module, we used our proposed fast approximate rank pooling to learn the motion of gestures, whereas a Gaussian guidance heatmap is learned in the static attention module under the train-phase-only supervision of skeleton data. 

%Meanwhile, the ACmW module is employed after each residual block to exploit the complementarity of input RGB and depth data. With a temporal convolutional layer, we combine the features of each frame by a simple 1-D convolutional layer. The entire network can be learned in an end-to-end manner.

\subsection{Local Network Structure Search in I3D }
\label{sec:net_search}

We use PC-darts \cite{xu2019pc}, which based on gradient descent and more  efficient than darts \cite{liu2018darts}, to search more efficient architecture for gesture recognition. In the search stage, we search and rebuild more adaptively structures and replace the Inception Modules in the I3D network with them. 
%while do not make any changes to the other layers. 
As shown in Fig.\ref{fig:pipeline}, because the features in different stages of the backbone have different solutions, we design three kinds of cells to learn the different levels of features. Specifically, 
we replace the first two Inception Modules in the I3D with \textbf{\textit{cell1}}, the middle five Inception Modules with \textbf{\textit{cell2}}, and the last two Inception Models with \textbf{\textit{cell3}}. %, where cell1, cell2 and cell3 are all different structures since we consider that the features captured by the network at different depths are different, and higher-level semantic features can often be captured in the deep layers of the network, so the shallow cell structures are no longer suitable for these layers. 
% Similar to PC-darts~\cite{xu2019pc}, 
Each cell represents a directed acyclic graph (DAG) with $k$ nodes $\{ m\}_{i=0}^{k-1}$. Each node indicates an output of a network layer, and each edge $(i, j)$ of the DAG indicates the information flow from node $m_i$ to $m_j$, which consists of the candidate operations weighted by the architecture parameter $\alpha^{(i,j)}$. 
Different from the I3D network which only employ  three operation '$Conv\_1 \times 1 \times 1$', '$Conv\_3 \times 3 \times3$' and  '$Max\_pooling\_3\times3\times3$'  in the original  Inception Module, we add two extra operations: '$Conv\_1 \times 3 \times3$' and '$Conv\_3 \times 1 \times1$'  to perform convolution either in spatial or temporal domain only.
That is because the size of features in the spatial domain (the height and width of frame) and temporal domain (the number of frames) differ a lot. Consequently, it is not necessary to perform the spatial and temporal convolution together all the time. Besides, a novel operator, dilated convolution '$dil\_3\times3\times3$', is introduced into search space for searching more powerful architecture.
%since sometimes the information of one of these two domains need preserving to the next layer. 
So the final search space $\mathcal{O}$ we defined includes seven candidate operations: '$Zero$', '$Identity$', '$dil\_3\times3\times3$', '$Conv\_1 \times 1 \times 1$', '$Conv\_3 \times 3 \times3$', '$Conv\_1 \times 3 \times3$', '$Conv\_3 \times 1 \times1$', where '$Zero$' and '$Identity$'mean no feature flow connection, direct feature flow without any convolution/pooling operations, respectively. '$Conv\_x\times y\times z$' and  '$dil\_x\times y\times z$' represent 3D vanilla and dilated convolution that kernel with the size of $x\times y\times z$ , respectively. Similar to the work of \cite{xu2019pc}, the architecture parameters $\alpha^{(i,j)}$ are optimized via the stochastic mini-batch gradient descent algorithm.
%Similar to the work of \cite{liu2018darts,xu2019pc}, when searching the network, we divide the training set into two parts with the ratio of $7:3$, {\color{red}{which are used to train the neural network and optimize the architecture parameters $\alpha^{(i,j)}$ through the gradient descent algorithm respectively.}} 
Specially, for each edge $(i,j)$, we can formulate it by a function $o'^{(i,j)}(\cdot)$ where  $o'^{(i,j)}(m_i) = \sum_{o \in \mathcal{O}} \eta_{o}^{(i,j)} ~\cdot ~ o(m_i)$ . Softmax $\eta_{o}^{(i,j)} =\sum_{o\in \mathcal{O} } \frac{exp(\alpha_{o}^{(i,j)})}{\sum_{{o}''\in \mathcal{O}}exp(\alpha_{{o}''}^{(i,j)})}$ is utilized to relax architecture parameter $\alpha^{(i,j)}$   into operation weight $o \in \mathcal{O}$. The intermediate node can be denoted as $m_j = \sum_{i<j}~o'^{(i,j)}(m_i)$. And the output node $m_{k-1}$  concat all the intermediate nodes.  The cross-entropy loss is utilized for the training loss $\mathcal{L}_{train}$ and validation loss $\mathcal{L}_{val}$. Then the network parameters $w$ and architecture parameters $\alpha^{(i,j)}$ are learned via solving the bi-level optimization problem:

\begin{equation}\label{eq:nas}
\begin{split}
&\mathop{min}\limits_{\alpha }\quad \mathcal{L}_{val} (w^{*}(\alpha), \alpha), \\
&s.t. \quad w^{*}(\alpha)=arg ~ min_{w}~ \mathcal{L}_{train}(w, \alpha)
\end{split}
\end{equation}

When the search converging, the optimal operation between the pair of node $(i,j)$  can be obtained by replacing each mixed operation $o'^{(i,j)}$  with the most likely operation: $o^{(i,j)}=\arg\max_{o \in \mathcal{O}, o \neq zero} ~\alpha_{o}^{(i,j)}$.

% the optimal operation between the pair of node $(i,j)$ can be obtained by  $max_{o \in \mathcal{O},o\neq zero} ~\eta_{o}^{(i,j)} $, where $o^{(i,j)}=\arg\max_{o \in \mathcal{O}, o \neq zero} ~\eta_{o}^{(i,j)}$.

\subsection{Dynamic-static Attention}
In the DSA module, two kinds of attention are learned by two sub-modules of dynamic attention and static attention, respectively. In the dynamic attention module, we used our proposed fast approximate rank pooling to learn the motion of gestures, whereas a Gaussian guidance heatmap is learned in the static attention module under the train-phase-only supervision of skeleton data. 
\subsubsection{\textit{Dynamic Attention Sub-module}}
The dynamic attention sub-module concerns the effective motion information among frames. We aggregate the intermediate spatiotemporal-structural information into a dynamic image instead of using the 3D manipulation directly like \cite{tran2015learning} to avoid time-consuming processing. To improve efficiency, we propose a fast approximate rank pooling algorithm based on \cite{bilen2016dynamic}, and it can reduces time complexity drastically.
{\flushleft \textit{Dynamic Image via Fast Approximate Rank Pooling.}}

According to \cite{smola2004tutorial}, the rank pooling map can be obtained via solving a convex optimization problem using the objective function of RankSVM. This process is known as rank pooling~\cite{fernando2017rank}. 
According to the work\cite{bilen2017action}, the approximation of rank pooling can be defined as:

\begin{equation}\label{eq:arp}
\mathbf{d}^{*} \propto \sum\limits_{t_1>t_2}{V_{t_1}-V_{t_2}}=\sum\limits_{t=1}^T{\beta_t V_t},
\end{equation}
where $T$ is the length of the video clip and $\beta_t=2t-T-1$. $V_t$ is the feature map of time step $t$. The result of Eq.(\ref{eq:arp}) is the dynamic image $DI$. In this computation, the time complexity is related to the number of frames, for a $T$-frame video clip, the time consumption can be up to $T(T-1)...1=T!$, and it is still high when processing a long-term video.

%In \cite{bilen2016dynamic}, Bilen \etal obtain the rank pooling map
\iffalse
The rank pooling map can be obtained via solving a convex optimization problem using the objective function of RankSVM~\cite{smola2004tutorial} in Eq.(\ref{eq:rp}):
\begin{equation}\label{eq:rp}
\begin{split}
\mathbf{d^{*}}=&\underset{\mathbf{d}}{\operatorname{argmin}} \{
\frac{\lambda}{2} \| \mathbf{d} \|^2 +  \frac{2}{T(T-1)} \times \\
 &\sum_{q>t}\operatorname{max}\{ 0, 1-S(q, \mathbf{d}) +S(t, \mathbf{d}) \} \},
\end{split}
\end{equation}
where $T$ is the video length, the first term is the parameter regularizer in SVM and the second term refers to how many incorrect pairs ranked by the ranking function $S(t,\mathbf{d})$. {\color{blue}{$\forall t_1 > t_2$, if we have $S(t_1, \mathbf{d}) > S(t_2, \mathbf{d})+1$, this pair is considered correctly.}} This process is known as rank pooling~\cite{fernando2017rank}. 
\fi

To simplify the processing of approximating rank pooling, we further study Eq.(\ref{eq:arp}), and achieve the fast approximate rank pooling via computing the relation between the dynamic image of frame $n$ to $m (m>n)$ and that of frame $n+1$ to $m+1$ as:

%When computing the dynamic image between frame $n$ and $m (m>n)$, 
%% according Eq.(\ref{eq:arp}) we have:
%{\color{red}{we can expand Eq.(\ref{eq:arp})  as bellow equation:}}
%\begin{small}
%\begin{equation}\label{eq:DIn}
%\begin{split}
%DI(n,m)& = V(I_{n+1}-I_{n}) \\
%       & + V(I_{n+2}-I_{n}) + V(I_{n+2}-I_{n+1})  \\
%       & ... \\
%       & + V(I_{m}-I_{n}) + V(I_{m}-I_{n+1}) + ... + V(I_{m}-I_{m-1})
%\end{split}
%\end{equation}
%\end{small}

%Similarly, the dynamic image between frame $n+1$ and $m+1$ can be:
%\begin{small}
%\begin{equation}\label{eq:DIn+1}
%\begin{split}
%DI(n+1,m+1)& = V(I_{n+2}-I_{n+1}) \\
%       & + V(I_{n+3}-I_{n+1}) + V(I_{n+3}-I_{n+2})\\
%       & \dots \\
%       & + V(I_{m+1}-I_{n+1}) + V(I_{m+1}-I_{n+2}) + \\
%       & \dots + V(I_{m+1}-I_{m})
%       %\vspace{-0.3cm}
%\end{split}
%\end{equation}
%\end{small}
%%\vspace{-0.3cm}
%Combining Eq.(\ref{eq:DIn}) and Eq.(\ref{eq:DIn+1}), we have:
%\begin{small}
%\begin{equation}\label{eq:DIcombine}
%\begin{split}
%DI(n+1,m+1)& = DI(n,m) \\
%       & - V(I_{n+2}-I_{n}) + V(I_{n+2}-I_{n}) + \\
%       & \dots + V(I_{m}-I_{n}) \\
%       & + V(I_{m+1}-I_{n+1}) + V(I_{m+1}-I_{n+2}) + \\
%       &\dots + V(I_{m+1}-I_{m}),
%\end{split}
%\end{equation}
%\end{small}
% Sort Eq.(\ref{eq:DIcombine}), 
%Finally, we can derive the computation of $DI(n+1,m+1)$ as:
\begin{equation}\label{eq:DInotime}
\begin{split}
DI(n+1,m+1)=&DI(n,m)+(m-n)\times V(I_n)+ \\
            &V(I_{m+1})-2\sum\limits_{l=n+1}^m V(I_l)
\end{split}
\end{equation}
where $I_n$ is the first frame of last dynamic image, and $I_{m+1}$ is the last frame of the current dynamic image. The last term is the overlapped part between two dynamic images. 
%(The details of the derivation can be seen in the supplementary material.) 
In this way, the computation of dynamic image can be irrelevant to the number of frames.

After obtaining the dynamic image, at each time $t$, we do the normalization for the rank parameters as:
\begin{equation}\label{eq:rp}
DI_{\{c,x,y\}} = \frac{DI_{\{c,x,y\}}-DI_{\rm{min}}}{DI_{\rm{max}}-DI_{\rm{min}}},
\end{equation}
where $DI_{(c,x,y)}$ represents the dynamic image at $c$-th channel and the coordinate of $(x,y)$. $DI_{\rm{min}}$ and $DI_{\rm{max}}$ are the minimum and maximum of the dynamic image, respectively.

% \begin{figure}[!hbp]
% \centering
% \subfigure[RGB frame]{\includegraphics[width=0.32\linewidth]{img/RGB_1.png}}
% \subfigure[dynamic attention map]{\includegraphics[width=0.32\linewidth]{img/RP_1.jpg}}
% \subfigure[static attention map]{\includegraphics[width=0.325\linewidth]{img/heatmap1.jpg}}
% \caption{An example of Gaussian skeleton map and dynamic image. (a) Input RGB image. (b) dynamic image. (c) the static attention map learned from the Gaussian skeleton map.}
% \label{fig:sample1}
% \end{figure}

% As shown in Fig.\ref{fig:sample1}(b), the dynamic image captures the movement of the performer's hands since the pixel of the image is changing in the temporal dimension. 
\begin{figure}[ht]
	\centering
	\includegraphics[width=0.9\linewidth]{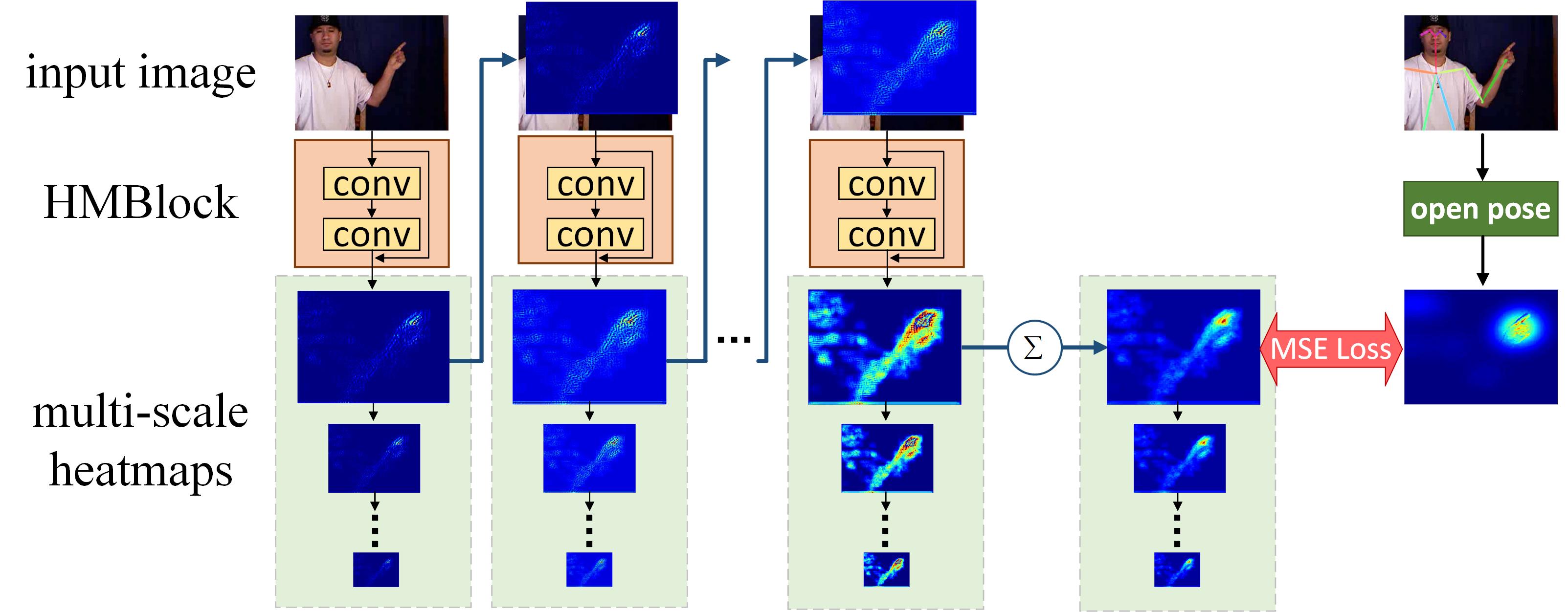}
	\caption{The framework of our HeatmapNet used for guidance heatmap generation. With the Gaussian skeleton map derived via OpenPose \cite{cao2017realtime} the ground truth, we learn a heatmap to indicate the hand/arm regions via a lightweight sub-network, which is composed of several cascaded Heatmap Blocks. These blocks can learn a increasingly clear heatmap stage-by-stage. }
	\label{fig:HeatmapNet}
\end{figure}

{\flushleft \textit{Structure of Dynamic Attention Sub-module.}}  Considering the value of dynamic images varies along with the amplitude of movement, which may be out of the range $[0, 1]$. Therefore, we first apply the normalization to the dynamic images. The batch normalization \cite{ioffe2015batch} is conducted to eliminate the influence of distribution inside a mini-batch:
\begin{equation}
{DI_{\rm{norm}}(t)} = (\frac{{DI(t)} -E[{DI(t)}]}{\sqrt{Var[{DI(t)}]}}) \times \gamma + \beta,
\end{equation}
where the expectation $E$ and variance $Var$ are computed over the training data set so that this normalization can consider the data distribution in the mini-batch, $\gamma$ and $\beta$ are the scale and shift parameter of the Batch Normalization layer.
We add 1 to each pixel of the guidance map to avoid the zero-value attention map, which may lead a vanishing of feature maps.
Then we can get the DAtt guided feature map $O_{D,t}^{l}$ as:

\begin{equation}\label{eq:+1}
O_{D,t}^{l} = [(DI_{\rm{norm}}^{l} + 1) \odot f_t^{l}] \otimes f_t^{l}.
\end{equation}
where $O_{D,t}^l$ is the guided feature map of layer $l$ at time step $t$ and $DI_{\rm{norm}}^{l}$ is dynamic gesture feature map of layer $l$ at time step $t$. Then, on each channel of the feature map $f_t^l$ ,  we use the element-wise product operation  $\odot$ to generate the attention map. After that, $O_{D,t}^l$ can be derived by the $1\times 1$  convolution of the corresponding attention map and the feature map $f_t^l$ , which we  describe with $\otimes$ in Eq.(\ref{eq:+1}). Finally, the shape of $O_{D,t}^l$ is the same as $f_t^l$. 

\textit{\textbf{Static Attention Sub-module.}} \label{sec:SAtt}
Being aware of the location of hands and arms is important to avoid the interference by gesture-irrelevant factors. Therefore, 
% as show in Fig.\ref{fig:sample1}(c), 
we try to highlight the location of hands/arms in each frame via the static attention sub-module. It is guided by a heatmap, which is related to the location of keypoints in hands/arms regions. The heatmap is derived from a lightweight online gesture region heatmap generation network – HeatmapNet. 

\textit{\textbf{HeatmapNet.}} As shown in Fig.\ref{fig:HeatmapNet}, to derive the ground truth of hands/arms' location, 
% we first generate a hand/arm-centred heatmap from skeleton data. Inspired by~\cite{lin2018large,narayana2018gesture},
we first employ the OpenPose \cite{cao2017realtime} to generate skeleton data from raw RGB videos.  Then we obtain the Gaussian skeleton map according to Eq.(\ref{eq:gaussianmap}):
\begin{equation}\label{eq:gaussianmap}
\mathcal{H}_t = \mathcal{G}(I_t, P_t, \sigma),
\end{equation}
where $I_t$ and $P_t$ are the input frame and the skeleton points at time $t$, and its corresponding Gaussian map is represented as $\mathcal{H}_t$, which is generated via Gaussian function $\mathcal{G}$ with standard deviation $\sigma$.
After having the Gaussian map as the ground truth, the HeatmapNet is used to learn the static guidance heatmap. Considering the cascaded network is always used in high-level vision tasks \cite{newell2016stacked} to learn both global and local cues and refine the details of feature map, we also use a cascaded structure to predict the static guidance map. In each stage of the network, we have a heatmap block (HMBlock) for prediction. The prediction of static guidance map for time step $t$ at stage $s$ can be formulated as:

\begin{equation}\label{eq:skblock}
\tilde{h}_t^s =
\begin{cases}
\mathcal{F}_{\theta}(I_t) & s=1\\
\mathcal{F}_{\theta}(I_t, \tilde{h}_t^{s-1}) & s>1,
\end{cases}
\end{equation}
where $\mathcal{F}_{\theta}$ is the residual block with parameter $\theta$. $I_t$ is the input frame at time step $t$ and $\tilde{h}_t^{s-1}$ is the predicted map at the previous stage $s-1$. With the stage-wise learning of the heatmap, the regions related to the gesture can be highlighted more apparently, 
% which is shown in Fig.\ref{fig:sample1}(c). 
Finally, the guidance maps in all the stages are combined by averaging via Eq.(\ref{eq:average}): %5.50731776+e8  2.0212530432+e10
\begin{equation}\label{eq:average}
h_{t} = \frac{1}{S}\sum_{s=1}^{S}{\tilde{h}_{t}^{s}},
\end{equation}
where $S$ is the number of stages. To better guide the features of gesture in both low-level and high-level, we then generate multi-scale guidance maps by max pooling. The size of these heatmaps is in accord with those of corresponding feature maps. Then the static guidance maps are fed into different layers of the network to highlight the gesture-relevant regions. Here we use MSE loss $\mathcal{L}^{m}_{hm}$  to learn the guidance heatmap. It needs to be emphasized that unlike the previous methods  \cite{liu2017continuous,wang2017largei} based on the detection techniques (\ie, faster R-CNN \cite{ren2015faster}) through the training and test phase, the HeatmapNet is only learned in the training phase, and avoids employing extra input data or complex computation when recognizing gestures. We only use the original skeleton data generated via OpenPose algorithm \cite{cao2017realtime} in the training phase, and in the inference phase, we only need to input the original image without additional skeleton information. And our network can automatically focus on the performer's hands and arms. Meanwhile,  The FLOPs of our HMBlock is about $5.51\times10^8$, whereas that of faster R-CNN is  about 36 times of ours at $2.02\times10^{10}$. Therefore, the addition of HMBlock in the training phase will not increase the burden of learning too much. 

Through the SAtt sub-module, we can obtain the guided feature map $O_{S,t}^l$ as Eq.~(\ref{eq:ssatt}):

\begin{equation}\label{eq:ssatt}
O_{S, t}^{l}=[(h_t^{l}+1)] \odot f_t^{l}] \otimes f_t^{l},
\end{equation}
where $O_{S,t}^l$ is the guided feature map of layer $l$ at time step $t$ and $h_t^{l}$ is static gesture heatmaps of layer $l$ at time step $t$. $h_l^t$ is generated via the HeatmapNet. Similar to the dynamic attention sub-module, the processing of "adding 1" to each pixel and the process for shape consistency are also conducted before the element-wise multiplication.
% {\color{red}{ Then, on each channel of the feature map $f_t^l$ ,  we use the element-wise product operation  $\odot$ to generate the attention map. The reshaping processing of the attention map $h_t^l$ is always conducted to maintain the shape consistency with the feature$f_t^l$. After that, $O_{t}^l$ can be derived by the $1\times 1$  convolution of the corresponding attention map and the feature map $f_t^l$ , which we  describe with $\otimes$ in Eg.(\ref{eq:ssatt}). Finally, the shape of $O_{S,t}^l$ is the same as $f_t^l$.}}

\subsection{Network Training\label{sec:train}}
%The detail of its training process is described in Section \ref{sec:train}. 
The entire network can be trained in an end-to-end manner after overall network architecture is searched out. 
The main branch of gesture recognition with RGB/depth data is trained together with the branch of HeatmapNet.
%we train the network with single RGB or depth data. The parameter of it is learned by minimizing the cross-entropy loss $\mathcal{L}^{m}_{cls}$. Meanwhile, in order to make the network more focused on performer’s hands and arms, we employ a lightweight network - HeatmapNet, which can generate static guidance maps online. 
The parameter of the main branch is learned by minimizing the cross-entropy loss $\mathcal{L}^{m}_{cls}$. Meanwhile, the online heatmap sub-network is trained together to learn a guidance heatmap via the MSE loss $\mathcal{L}^{m}_{hm}$. Then we jointly optimize the entire network with a multi-task loss function. The Loss function can be expressed as:
\begin{equation}\label{eq:1st_loss}
\begin{split}
\mathcal{L}^{m} &= \mathcal{L}^{m}_{cls} + \gamma \mathcal{L}^{m}_{hm} \\
                &= -\sum\limits_{k=1}^K p^{m}_k log(\hat{p^{m}_k}) +
                \gamma ||\mathcal{F}_{\theta}(I^m)-\mathcal{H}||^2,
\end{split}
\end{equation}
where $m$ indicates the modality of data, which can be either RGB or depth. $p_k=\{p_1,p_2,\dots,p_K\}$ is the ground-truth probability distribution of the $k$-th class of the gesture, and $\hat{p_k}$ is its estimation. $\mathcal{H}$ is the ground truth Gaussian skeleton map, and $\mathcal{F}_{\theta}(\cdot)$ is the mapping function the sub-network learning for HMBlocks as mentioned in Section \ref{sec:SAtt}. $\gamma$ is the balancing parameter and we have $\gamma=100$ in this paper.

\section{Experiments}
\subsection{Datasets}
We evaluate our method and compare it with other state-of-the-art methods on two RGB-D gesture datasets:  Chalearn IsoGD dataset \cite{wan2016chalearn} and NvGesture dataset \cite{molchanov2016online}. Meanwhile, we also conduct the ablation studies on a hand-centred action dataset, THU-READ dataset \cite{tang2017action,tang2018multi} to show the generality of our network.

\subsection{Experimental Setup}
Our experiments are all conducted with Pytorch on the NVIDIA RTX 2080 Ti GPU. During the training process, the inputs are spatially resized to $256\times 256$ and then cropped into $224\times224$ randomly in the training stage, and are center cropped into $ 224\times 224$ in the test stage. The data is fed into the network with a mini-batch of 64 samples. For optimization, We use the SGD optimizer to train our network with the weight decay of 0.0003 and the momentum of 0.9. The learning rate is initially fixed as 0.01 and if the accuracy on the validation set not improved every 3 epochs, it is reduced by 10 times. The training work is stopped after 80 epochs or when the learning rate is under 1e-5. 
% The backbone network is chosen as I3D \cite{carreira2017quo} for Chalearn IsoGD dataset and NvGesture dataset. 
\subsection{Comparison with State-of-the-art Methods}
Our method is compared with recent state-of-the-art methods on IsoGD and NvGesture dataset. Table \ref{tab:IsoGD} shows the comparison on IsoGD dataset. Since most of methods release their result on the validation subset, we also conduct experiments on it for a fair comparison.

\begin{table}[ht]
\centering
\resizebox{1\linewidth}{!}{
\begin{tabular}{c c c}
\toprule
 Method & Modality & Accuracy (\%) \\
 %\hline
 \midrule
c-ConvNet \cite{wang2018cooperative}   & RGB &36.60\\
C3D-gesture  \etal \cite{li2017largeb} & RGB & 37.28  \\
AHL \cite{Hu2018Learning}  &RGB & 44.88 \\
ResC3D  \cite{miao2017multimodal} &RGB &45.07  \\
3DDSN \cite{duan2018unified} & RGB &46.08 \\
3DCNN+LSTM \cite{zhang2017learning}  & RGB &51.31 \\
attentionLSTM \cite{zhang2018attention}  & RGB & 55.98 \\
Redundancy+AttentionLSTM \cite{zhu2019redundancy}  & RGB & 57.42 \\
\textbf{RAAR3DNet(Ours)} &RGB &\textbf{62.66} \\ 
%\textbf{RAAR3DNet(Ours)} &RGB &\textbf{58.45} \\ %&\textbf{64.98\%}\\
\midrule
%Narayana   \etal \cite{narayana2018gesture} &Depth &27.98 \\
c-ConvNet  \cite{wang2018cooperative}    & Depth &40.08\\
C3D-gesture \cite{miao2017multimodal}  & Depth & 40.49  \\
ResC3D \cite{li2017largea}  &Depth &48.44 \\
AHL \cite{Hu2018Learning} &Depth & 48.96  \\
3DCNN+LSTM \cite{zhang2017learning} & Depth &49.81 \\
attentionLSTM \cite{zhang2018attention}   & Depth & 53.28 \\
Redundancy+AttentionLSTM \cite{zhu2019redundancy}  & Depth & 54.18 \\
3DDSN \cite{duan2018unified} & Depth &54.95 \\
  \textbf{RAAR3DNet(Ours)} &Depth &\textbf{60.66} \\
%  \textbf{RAAR3DNet(Ours)} &Depth &\textbf{57.34} \\
 \midrule
c-ConvNet  \cite{wang2018cooperative}    & RGB-D & 44.80\\
AHL \cite{Hu2018Learning}   &RGB-D & 54.14\\
3DCNN+LSTM \cite{zhang2017learning} &RGB-D & 55.29 \\
Redundancy+AttentionLSTM \cite{zhu2019redundancy}  & RGB-D & 61.05 \\
%{\color{red}{ResC3D \cite{miao2017multimodal} & RGB-D & }} \\
\textbf{RAAR3DNet(Ours)} & RGB-D &\textbf{66.62}  \\
% \textbf{RAAR3DNet(Ours)} & RGB-D &\textbf{63.53}  \\

 \bottomrule
\end{tabular}
}
\caption{Results on  IsoGD dataset. }
\label{tab:IsoGD}
\end{table}

As can be seen in Table \ref{tab:IsoGD}, our proposed method achieves the best performance on all conditions of using single RGB/depth data and using the fusion of them. For RGB data, our method outperforms the second-best one, \cite{zhu2019redundancy} at about $5\%$, even though their methods are trained with a complex combination of Res3D, Gated convLSTM and 2D CNNs. Ours is also $17\%$ higher than ResC3D \cite{li2017largea}, which achieves 1st place in the 2nd round of Chalearn LAP large-scale isolated gesture recognition challenge. The performance on the depth data is a little lower than that on RGB data. That may because the texture showing the details of fingers is not available in depth data. However, the recognition result on depth data is similar to that on RGB data. Ours also outperform the second-best one, 3DDSN \cite{duan2018unified} about $6\%$. The performance on the fusion RGB-D data also shows the effectiveness of our method, which outperforms Zhu \etal's method at about $5.6\%$.
%It worth pointing that although only using RGB and depth data for fusion, our methods still outperforms  other methods even employ other modalities of data for fusion, like optical flow {} and saliency {} data.

\begin{table}[ht]
\centering
\resizebox{1\linewidth}{!}{
\begin{tabular}{c c c}

 \toprule
 Method & Modality & Accuracy (\%) \\
 %\hline
 \midrule
HOG+HOG$^2$ \cite{ohn2014hand}  &RGB & 24.50 \\
Simonyan \etal \cite{simonyan2014two} &RGB & 54.60 \\
Wang \etal \cite{wang2016robust}  &RGB & 59.10 \\
C3D \cite{tran2015learning} &RGB & 69.30 \\
R3DCNN \cite{molchanov2016online}  &RGB & 74.10 \\
GPM \cite{gupta2019progression}  &RGB & 75.90 \\
PreRNN \cite{yang2018making}  &RGB & 76.50 \\
ResNeXt-101 \cite{kopuklu2019real} &RGB & 78.63 \\
MTUT \cite{abavisani2019improving}  &RGB & 81.33 \\

\textbf{RAAR3DNet(Ours)} &RGB &\textbf{85.83}\\
\midrule
HOG+HOG$^2$ \cite{ohn2014hand}  &Depth & 36.30 \\
SNV \cite{yang2014super}  &Depth &70.70 \\
C3D \cite{tran2015learning}  &Depth & 78.80 \\
R3DCNN \cite{molchanov2016online} &Depth & 80.30 \\
ResNeXt-101 \cite{kopuklu2019real}  &Depth & 83.82 \\
PreRNN \cite{yang2018making} &Depth & 84.40 \\
MTUT \cite{abavisani2019improving}  &Depth & 84.85 \\
GPM \cite{gupta2019progression}  &Depth & 85.50 \\

\textbf{RAAR3DNet(Ours)}  &Depth &\textbf{86.67}\\
\midrule
HOG+HOG$^2$ \cite{ohn2014hand} &RGB-D & 36.90 \\
I3D \cite{carreira2017quo} &RGB-D & 83.82 \\
PreRNN \cite{yang2018making} &RGB-D & 85.00 \\
MTUT \cite{abavisani2019improving}  &RGB-D & 85.48 \\
GPM \cite{gupta2019progression}  &RGB-D & 86.10 \\
\hline
human & &88.4 \\
\hline
\textbf{RAAR3DNet(Ours)}  &RGB-D &\textbf{88.59} \\
 \bottomrule
\end{tabular}
}
\caption{Results on the NvGesture dataset.}
\label{tab:NVGesture}
\end{table}

The comparison on NvGesture dataset is shown in Table \ref{tab:NVGesture}. As can be seen, our method can still achieve a competitive result on this dataset. Compared with the method of MTUT \etal \cite{molchanov2016online}, which uses a combination of time-consuming models of C3D and LSTM, our network achieves about $4.5\%$ improvement on RGB data. Meanwhile, it outperforms GPM \cite{gupta2019progression}, the second-best result on depth data at about $1.17\%$. The gap between the performance of ours and GPM becomes more significant on the fusion of RGB-D data at $2.49\%$. Noticing that our result is also better than the human recognition accuracy even without auxiliary data like IR as provided in that dataset.  It shows the effectiveness of our network architecture searching strategy and the attention module for improving the recognition performance.

\begin{figure}[ht]
\centering
\includegraphics[width=0.40\textwidth]{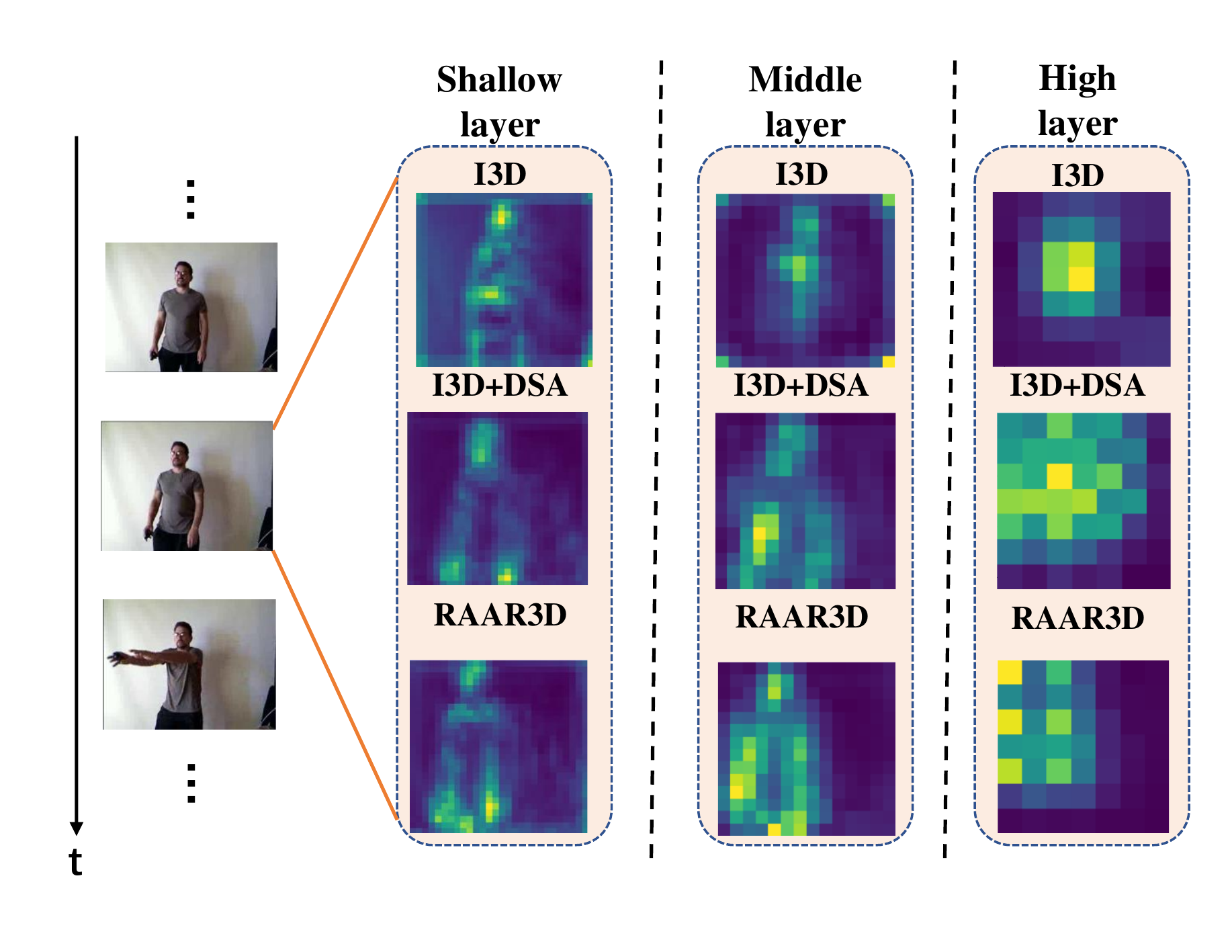}
\caption{Visualization of feature maps in the early, middle and late stage of the network. The feature maps from top to bottom are with the I3D, I3D with DSA module and our ultimate RAAR3D network, respectively.
}
\label{fig:featuremap}
\end{figure}
\subsection{Ablation Studies}
In this section, we perform several groups of experiments  to verify the effect of each component of our network, including the idea of automatically searching the architecture of network and the spatiotemporal DSA module.

\textit{\textbf{Effect of Network Architecture Searching.}} Here we first compare the performance of the baseline raw I3D network and our proposed  network on three different datasets - IsoGD, NvGesture and THU-READ dataset.

As shown in Table \ref{tab:i3d}, with a sophisticated network structure that searched via NAS, the performance on three datasets can be about $1\%$ higher than the original I3D network on RGB data and about $2\%$ higher on depth data. To better illustrate the effect of architecture searching, we also visualize the features of the original I3D network and those of the reconstructed network in Fig.\ref{fig:featuremap}. From the visualization result we can clearly find that via searching the optimal structure of network, the response of the background and other gesture-irrelevant regions in the feature maps is decreased significantly. It shows that searching and reconstructing the internal structure of a network architecture can make it more suitable for the specific task and extract more reasonable features. 

\textit{\textbf{Effect of Attention Mechanism.}} Table \ref{tab:att} shows the influence of  different attention schemes. For a fair comparison, we remove other modules and only take the original I3D network as the baseline. We give the result of the original I3D, the network only with dynamic attention or static attention, and finally the entire DSA module.

As shown in Table \ref{tab:att},  
\begin{table}[ht]
\centering
\resizebox{0.47\textwidth}{!}{
\begin{tabular}{c|c|c|c|c}
\hline

\hline
\multirow{2}{*}{\diagbox{Dataset}{Network}} &\multicolumn{2}{c|}{I3D} &\multicolumn{2}{c}{RAAR3DNet (w/o DSA)}  \\
 \cline{2-5} & RGB &Depth & RGB &Depth \\
\hline

% IsoGD           & 56.15\% & 54.64\% & \textbf{57.36}\% &\textbf{56.60} \% \\
IsoGD           & 59.15\% & 56.64\% & \textbf{61.13}\% &\textbf{58.32} \% \\
NvGesture  & 81.67\% &83.12\% & \textbf{82.50}\% &\textbf{85.83}\% \\
THU-READ & 65.50\% & 67.50\% &\textbf{67.92}\% & \textbf{68.75}\% \\
\hline

\hline
\end{tabular}
}
\caption{Comparisons with raw I3D and RAAR3DNet on the Gesture and Action dataset.}
\label{tab:i3d}
\end{table}
the DSA module can improve the performance at about 3\% on the basic I3D network, which implies the attention mechanism plays a very important role in the video-based recognition tasks since it can help the network to focus on the most noteworthy regions. Compared with the dynamic attention sub-module, the gain brought by static attention is a little higher. The reason for it may be the details of hands/arms is more important for recognizing some subtle differences. Therefore, the highlight of regions of hands/arms contributes more to the improvement.

\section{Conclusion}
In this paper, we propose a regional attention with searched architecture 3D network for gesture recognition basing on RGB-D data. We take the I3D network as the backbone, and employ NAS to search the optimal connection among features in different stages of it. In this way, the structure of the network can fit the low-level and high-level features better and improves the recognition result. Meanwhile, we also design a stackable attention module of DSA to guide the network to pay more attention to the hand/arm regions in each frame and the motion trajectory among video sequence. Finally, comparisons with state-of-the-art methods on two gesture datasets prove the effectiveness of the proposed method.
\begin{table}[ht]
\centering
\begin{threeparttable}
\begin{tabular}{c|c|c}
\hline
\multirow{2}{*}{Strategy} &\multicolumn{2}{c}{Recognition Rate}  \\
 \cline{2-3} & RGB &Depth \\
\hline
baseline(I3D) &81.67\% &83.12\% \\
DAtt &83.54\% &84.58\%  \\
SAtt &83.75\% &85.21\% \\
DSA(DAtt+SAtt) &\textbf{84.79\%} &\textbf{85.83\%} \\
\hline
\end{tabular}
\end{threeparttable}
% \begin{tablenotes}
% \item[*] \small {DAtt = dynamic attention,  SAtt = static attention}
% \end{tablenotes}
\caption{Performance of attention modules on the NVGesture dataset. Where DAtt and SAtt indicate the dynamic attention module and static attention module, respectively.}
\label{tab:att}
\end{table}
\section{Acknowledgments}
This work was supported by the Chinese National Natural Science Foundation Projects $\#$61961160704, $\#$61876179, $\#$62002271, $\#$61772396, $\#$61772392, $\#$61902296, the External cooperation key project of Chinese Academy Sciences $\#$ 173211KYSB20200002, the Key Project of the General Logistics Department Grant No.AWS17J001, Science and Technology Development Fund of Macau (No.~0010/2019/AFJ, 0025/2019/AKP), the National Key R\&D Program of China under Grant $\#$2018YFC0807500, the Fundamental Research Funds for the Central Universities $\#$JBF180301, Xi’an Key Laboratory of Big Data and Intelligent Vision $\#$201805053ZD4CG37.

\bibliography{Formatting-Instructions-LaTeX-2021}
\end{document}

\def\etal{\emph{et al}.}
\def\eg{\emph{e}.\emph{g}.}
\def\ie{\emph{i}.\emph{e}.}

\maketitle

\section{Overview}
In this supplementary material, we provide a more detailed analysis of our methods and experiments that are not shown in the main paper. we  ﬁrst present the details of the proposed attention modules, including the derivation of the fast approxmiate rank pooling and the operations in each block of SAtt and DAtt,  visualize the neural activation maps of them. Finally, we compared the neural activation maps of the I3D and RAAR3DNet networks on the random 10 samples of the NV-Gesture test dataset to further prove the effectiveness of the method that automatically rebuilt structure through the network via NAS. 

\section{Dynamic-static Attention }
%  (Sec. 3.1 in the manuscript)
\subsection{\textit{Details of the DAtt sub-module}} \label{datt}
The details of the derivation of Eq.4, which are not given in the manuscript are as follow:

To compute the dynamic image between frame $n$ and $m (m>n)$, according Eq.3 in the manuscript,  we have:
\begin{small}
\begin{equation}\label{Eq:DIn}
\begin{split}
DI(n,m)& = V(I_{n+1}-I_{n}) \\
       & + V(I_{n+2}-I_{n}) + V(I_{n+2}-I_{n+1})  \\
       & ... \\
       & + V(I_{m}-I_{n}) + V(I_{m}-I_{n+1}) + ... + V(I_{m}-I_{m-1})
\end{split}
\end{equation}
\end{small}

Similarly, the dynamic image between frame $n+1$ and $m+1$ can be:
\begin{small}
\begin{equation}\label{Eq:DIn+1}
\begin{split}
DI(n+1,m+1)& = V(I_{n+2}-I_{n+1}) \\
       & + V(I_{n+3}-I_{n+1}) + V(I_{n+3}-I_{n+2})\\
       & \dots \\
       & + V(I_{m+1}-I_{n+1}) + V(I_{m+1}-I_{n+2}) + \\
       & \dots + V(I_{m+1}-I_{m})
       %\vspace{-0.3cm}
\end{split}
\end{equation}
\end{small}
%\vspace{-0.3cm}
Combine Eq.(\ref{Eq:DIn}) and Eq.(\ref{Eq:DIn+1}), we have:
\begin{small}
\begin{equation}\label{Eq:DIcombine}
\begin{split}
DI(n+1,m+1)& = DI(n,m) \\
       & - V(I_{n+2}-I_{n}) + V(I_{n+2}-I_{n}) + \\
       & \dots + V(I_{m}-I_{n}) \\
       & + V(I_{m+1}-I_{n+1}) + V(I_{m+1}-I_{n+2}) + \\
       &\dots + V(I_{m+1}-I_{m}),
\end{split}
\end{equation}
\end{small}

Sort Eq.(\ref{Eq:DIcombine}), then we can derive the computation of $DI(n+1,m+1)$ as:
\begin{equation}\label{eq:DInotime}
\begin{split}
DI(n+1,m+1)=&DI(n,m)+(m-n)\times V(I_n)+ \\
            &V(I_{m+1}))-2\sum\limits_{l=n+1}^m V(I_l)
\end{split}
\end{equation}
\begin{figure}[!h]
\centering
\subfigure[RGB frame]{\includegraphics[width=0.32\linewidth]{img/RGB_1.png}}
\subfigure[DAtt map]{\includegraphics[width=0.32\linewidth]{img/RP_1.jpg}}
\subfigure[SAtt map]{\includegraphics[width=0.325\linewidth]{img/heatmap1.jpg}}
\caption{An example of dynamic image and gaussian skeleton map. (a) Input RGB image; (b) Dynamic image; (c) The static attention map learned from the gaussian skeleton map.}
\label{fig:sample1}
\end{figure}

The size of features and attention maps in the DAtt is shown in Table \ref{tab: DAtt}. The feature $f_t^l$ derived from the previous layer is with the shape of $N \times C \times T \times H \times W$, which are the number of batch size, channel, depth (\ie, the number of frames for the video), height and weight of the feature map, respectively. For each stage of the network, in order to eliminate the inﬂuence of distribution inside a mini-batch, we first apply the normalization to the dynamic images. Take the first block as an example, we perform batch normalization \cite{ioffe2015batch} on input images, and get the output $DI_t^{1}$ with the dimension of $N \times 192 \times 16 \times 28 \times 28$ for batch size, channel, depth, height and weight.  And then we use some basic operations (\ie, Adding 1 to each pixel of the guidance map, element-wise product operation and convolution operation) to achieve the dynamic attention and obtain the output feature map $O_{D,t}^{1}$ with the size of $N \times 192 \times 16 \times 28 \times 28$.  For the second to three stages, we did the same operation to fuse the two features, the only difference is that the dimensions of the input SAtt sub-module are different at each stage.

As shown in Fig.\ref{fig:sample1}(b), the dynamic image captures the movement of the performer's hands/arms since the pixel of the image is changing in the temporal dimension. 

\begin{table}[!h]
\centering
\scalebox{0.8}{
\begin{tabular}{c|c|c|c|c}
\hline

\hline
  Module & & &Layer & Output Size  \\
\hline

\hline
  \multirow{24}{*}{\textbf{DAtt}}& \multirow{7}{*}{stage1} &  $f_t^{1}$& Input 
   &  $N \times 192 \times 16 \times 28 \times 28$\\ \cline{3-5} 
%    &&\multirow{3}{*}{$DI_t^{1}$ } & Padding & $6\times64 \times 56 \times 56 $ \\
%    &&&Rankpooling & $6\times 64 \times 56 \times 56 $ \\
% &&& Out & $6\times64 \times 56 \times 56 \times 16 $ \\ \cline{3-5} 
&&\multirow{6}{*}{Block1 } &$DI_t^{1}$&  $N \times 192 \times 16 \times 28 \times 28$\\
&&&  BatchNorm2d & $N \times 192 \times 16 \times 28 \times 28$  \\
&&&Add 1 &  $N \times 192 \times 16 \times 28 \times 28$ \\
&&& $\odot$  & $N \times 192 \times 16 \times 28 \times 28$ \\
&&& $\otimes$  &  $N \times 192 \times 16 \times 28 \times 28$  \\
&&& Output  &  $N \times 192 \times 16 \times 28 \times 28$ \\ \cline{2-5} 

& \multirow{7}{*}{stage2} &  $f_t^{2}$& Input 
   & $N\times 256 \times 8 \times 14 \times 14$\\ \cline{3-5} 
%    &&\multirow{3}{*}{$DI_t^{2}$ } & Padding & $6\times 128 \times 28 \times 28 $ \\
%    &&&Rankpooling & $6\times 128 \times 28 \times 28 $ \\
% &&& Out & $6\times 128 \times 28 \times 28 \times 16 $ \\ \cline{3-5} 
&&\multirow{6}{*}{Block2 } &$DI_t^{2}$ & $N\times 256 \times 8 \times 14 \times 14$ \\
&&&BatchNorm2d &$N\times 256 \times 8 \times 14 \times 14$  \\
&&&Add 1 & $N\times 256 \times 8 \times 14 \times 14$ \\
&&& $\odot$  &$N\times 256 \times 8 \times 14 \times 14$ \\
&&& $\otimes$  & $N\times 256 \times 8 \times 14 \times 14$  \\
&&& Output  & $N\times 256 \times 8 \times 14 \times 14$ \\ \cline{2-5} 

& \multirow{7}{*}{stage3} &  $f_t^{3}$& Input 
   & $N\times 512 \times 4 \times7 \times 7$\\ \cline{3-5} 
%    &&\multirow{3}{*}{$DI_t^{3}$ } & Padding & $6\times 256 \times 14 \times 14 $ \\
%    &&&Rankpooling & $6\times 256 \times 14 \times 14 $ \\
% &&& Out & $6\times 256 \times 14 \times 14 \times 16 $\\ \cline{3-5} 
&&\multirow{6}{*}{Block3 } &  $DI_t^{3}$ &$N\times 512 \times 4 \times7 \times 7$\\
&&&BatchNorm2d &$N\times 512 \times 4 \times7 \times 7$ \\
&&&Add 1 & $N\times 512 \times 4 \times7 \times 7$ \\
&&& $\odot$  &$N\times 512 \times 4 \times7 \times 7$ \\
&&& $\otimes$  & $N\times 512 \times 4 \times7 \times 7$  \\
&&& Output  & $N\times 512 \times 4 \times7 \times 7$ \\ \cline{2-5} 

% & \multirow{7}{*}{stage4} &  $f_t^{4}$& Input 
%    & $N\times 512 \times 7 \times7 \times 16$\\ \cline{3-5} 
% %    &&\multirow{3}{*}{$DI_t^{4}$ } & Padding & $6\times 512 \times 7 \times 7 $ \\
% %    &&&Rankpooling & $6\times 512 \times 7 \times 7 $ \\
% % &&& Out & $6\times 512 \times 7 \times 7 \times 16 $\\ \cline{3-5} 
% &&\multirow{6}{*}{Block4 } & $DI_t^{4}$ &$N\times 512 \times 7 \times 7 \times 16 $\\
% &&&BatchNorm2d &$N\times 512 \times 7 \times 7\times 16 $ \\
% &&&Add 1 & $N\times 512 \times 7 \times 7 \times 16$ \\
% &&& $\odot$  &$N\times 512 \times 7 \times 7 \times 16$ \\
% &&& $\otimes$  & $N\times 512 \times 7 \times 7\times 16 $  \\
% &&& Output  & $N\times 512 \times 7 \times 7 \times 16 $ \\ \cline{2-5} 
\hline

\hline
 \end{tabular}
 }
 \caption{Details about the features size in the DAtt sub-module, which is used to capture the motion information among frames. DAtt is embedded in the RAAR3DNet in three stages. }
 \label{tab: DAtt}
\end{table}

\subsection{\textit{Details of the SAtt sub-module}} \label{satt}
The size of features and attention maps in the SAtt is shown in Table \ref{tab: SAtt}. The features $f_t^l$ from previous DAtt sub-module are with the different shapes in different stage.
For each stage of the network, we first generate the static guidance maps through our lightweight network - HeatmapNet, and its dimension is $N \times 1 \times 224 \times 224$ for batch size, channel, height and weight. Take the block 1 as an example, to make its dimension consistent with the dimensions of features $f_t^l$, we change its height and weight to $28$ by max-pooling and extend its depth to $16$ by concatenating the attention map $16$ times. Then we obtain the guidance map $h_t^l$. Owing to the same size between $h_t^l$ and $f_t^l$, similar to DAtt, we also use some basic operations (\ie, Adding 1 to each pixel of the guidance map, element-wise product operation and convolution operation ) to achieve the static attention and obtain the output feature map $O_t^l$ with the size of $N \times 192 \times 16 \times 28 \times 28$. Then the $O_t^l$ is taken as the input as the succeed next layer's input. 
For the second to three stages, we did the same operation to fuse the two features, the only difference is that the dimensions of the input next layer are different at each stage.

As shown in Fig.\ref{fig:sample1}(c), the static attention mainly helps outline the region of performer‘s hands and arms.

\begin{table}[!h]
\centering

\scalebox{0.8}{
\begin{tabular}{c|c|c|c|c}
\hline

\hline
  Module & & &Layer & Output Size  \\
\hline

\hline
  \multirow{20}{*}{\textbf{SAtt}}& \multirow{6}{*}{stage1} &  $f_t^{1}$& Input &   $N \times 192 \times 16 \times 28 \times 28$ \\ \cline{3-5} 
% &&\multirow{6}{*}{$h_t^{1}$ } & Conv$3\times1$ & $6\times3 \times 224 \times 224 $ \\
% &&& BatchNorm2d & $6\times 3 \times 224 \times 224 $ \\
% &&& ReLU & $6\times 3 \times 224 \times 224 $ \\
% &&& Conv$3\times1$  & $6\times1 \times 224 \times 224 $ \\
% &&&BatchNorm2d  & $6\times 1 \times 224 \times 224 $ \\
% &&&Out  & $6\times 1 \times 224 \times 224 \times 16 $ \\\cline{3-5} 
&&\multirow{5}{*}{Block1 } &  $h_t^{1}$ &  $N \times 192 \times 16 \times 28 \times 28$ \\
&&&Add 1 &   $N \times 192 \times 16 \times 28 \times 28$ \\
&&& $\odot$  &   $N \times 192 \times 16 \times 28 \times 28$ \\
&&& $\otimes$  &  $N \times 192 \times 16 \times 28 \times 28$ \\ 
&&& Output  &   $N \times 192 \times 16 \times 28 \times 28$ \\ \cline{2-5} 

& \multirow{6}{*}{stage2} & $f_t^{2}$ & Input & $N\times 256 \times 8 \times 14 \times 14$ \\ \cline{3-5} 
% &&\multirow{6}{*}{$h_t^{2}$ } & Conv$4\times1$ & $6\times4 \times 224 \times 224 $ \\
% &&& BatchNorm2d & $6\times 4 \times 224 \times 224$ \\
% &&& ReLU & $6\times 4 \times 224 \times 224 $ \\
% &&& Conv$4\times1$  & $6\times1 \times 224 \times 224$ \\
% &&&BatchNorm2d  & $6\times 1 \times 224 \times 224 $ \\ 
% &&&Out  & $6\times 1 \times 224 \times 224 \times 16 $ \\\cline{3-5} 
&&\multirow{5}{*}{Block2 } &  $h_t^{2}$ &  $N\times 256 \times 8 \times 14 \times 14$ \\
&&&Add 1 &  $N\times 256 \times 8 \times 14 \times 14$ \\
&&& $\odot$  &  $N\times 256 \times 8 \times 14 \times 14$ \\
&&& $\otimes$  & $N\times 256 \times 8 \times 14 \times 14$ \\
&&& Output  &  $N\times 256 \times 8 \times 14 \times 14$ \\ \cline{2-5} 

& \multirow{6}{*}{stage3} & $f_t^{3}$ & Input & $N\times 512 \times 4 \times7 \times 7$ \\ \cline{3-5} 
% &&\multirow{6}{*}{$h_t^{3}$ } & Conv$4\times1$ & $6\times4 \times 224 \times 224 $ \\
% &&& BatchNorm2d & $6\times 4 \times 224 \times 224 $ \\
% &&& ReLU & $6\times 4 \times 224 \times 224 $ \\
% &&& Conv$4\times1$  & $6\times1 \times 224 \times 224 $ \\
% &&&BatchNorm2d  & $6\times 1 \times 224 \times 224 $ \\
% &&&Out  & $6\times 1 \times 224 \times 224 \times 16 $ \\\cline{3-5} 
&&\multirow{4}{*}{Block3 } &  $h_t^{3}$ & $N\times 512 \times 4 \times7 \times 7$ \\
&&&Add 1 & $N\times 512 \times 4 \times7 \times 7$ \\
&&& $\odot$  & $N\times 512 \times 4 \times7 \times 7$ \\
&&& $\otimes$  &$N\times 512 \times 4 \times7 \times 7$ \\
&&& Output  & $N\times 512 \times 4 \times7 \times 7$ \\ \cline{2-5} 

% & \multirow{6}{*}{stage4} & $f_t^{4}$ & Input & $N\times 512 \times7 \times 7 \times 16$ \\ \cline{3-5} 
% % &&\multirow{6}{*}{$h_t^{4}$ } & Conv$4\times1$ & $6\times4 \times 224 \times 224 $ \\
% % &&& BatchNorm2d & $6\times 4 \times 224 \times 224 $ \\
% % &&& ReLU & $6\times 4 \times 224 \times 224 $ \\
% % &&& Conv$4\times1$  & $6\times1 \times 224 \times 224 $ \\
% % &&&BatchNorm2d  & $6\times 1 \times 224 \times 224 $ \\
% % &&&Out  & $6\times 1 \times 224 \times 224 \times 16 $ \\\cline{3-5} 
% &&\multirow{4}{*}{Block4 } &  $h_t^{4}$ & $N\times 512 \times 7 \times 7\times 16$ \\
% &&&Add 1 & $N\times 512 \times 7 \times 7\times 16$\\
% &&& $\odot$  & $N\times 512 \times 7 \times 7\times 16$\\
% &&& $\otimes$  & $N\times 512 \times 7 \times 7\times 16$ \\
% &&& Output  & $N \times 512 \times 7 \times 7 \times 16$ \\ \cline{2-5} 
\hline

\hline
 \end{tabular}
 }
 \caption{Details about the size of features in the SAtt sub-module, which is used to indicate the location of hands and arms in a video frame. Along with the three groups of blocks in the RAAR3DNet, the SAtt sub-module also has three stages, and the shape of data changes in different stages. }
\label{tab: SAtt}

\end{table}

% \subsection{\textit{Details of the HeatmapNet Architecture}} \label{heatmapNet}

\begin{figure*}[!h]
\centering
\scalebox{0.9}{
\includegraphics[width=0.9\linewidth]{img/visualize_feature.png}
}
\caption{Feature visualization from I3D assembled with varied attention modules on the NV-Gesture dataset. The last four rows represent the neural activation with  I3D, DAtt, SAtt, and DSA, respectively.}
\label{fig:vis_feature}
\end{figure*}
\section{Visualization and Analysis}\label{sec:heatmap}}
% In this section, we visualize the feature response of 
\subsection{\textit{Feature Visualization of the Attention Module}}
% \begin{figure*}[!h]
% \centering
% \includegraphics[width=1.0\linewidth]{img/SAtt-DAtt_New1.png}

% \caption{Visualization of intermediate feature maps of SAtt and DAtt sub-modules.  We randomly selected a sample and visualized the feature maps extracted by these two sub-modules at different stages.}
% \label{fig:satt_datt}
% \end{figure*}
The  feature maps of our attention modules are visualized in Fig.\ref{fig:vis_feature}. Comparing with the original I3D, it can be seen that the proposed DAtt and SAtt sub-modules can significantly enhance the spatial and temporal representation, respectively. And our DSA module can finally fuses these two kinds of attention to drive model to focus more on the trajectories arms and hands. Specifically,  we find SAtt mainly helps outline the region of performers. Compared with the feature maps without any attention mechanism, the features of SAtt can steadily depict where the body, the arm and the hand of the performer is. The DAtt features, on the other hand, are less related to the outline of performer but closely tied to the intensity of movement. Then the movement paths are highlighted in the DAtt features. In addition, integrating the advantages of SAtt and DAtt, the DSA can have a very significant effect on the appearance of feature maps, which indicates the contextual information of the movement path. The DSA module not only marks the regions related to the gesture, like the arm of the performer, but also distinguishes the ranges of motion at different positions of the video sequence. It effectively avoids the impact of gesture-irrelevant regions, which have a high response in the features without attention mechanism, especially when there is a drastic movement like raising or dropping the arm. Therefore, our DSA module can better guide the network to focus on the hands/arms and give a more accurate prediction. 

\begin{figure*}[!h]
\centering
\scalebox{1.0}{
\subfigure[]{\includegraphics[width=0.45\linewidth]{img/I3D_feature1.png}}
\subfigure[]{\includegraphics[width=0.45\linewidth]{img/Aut_features1.png}}
}
\caption{Feature response of the random 10 samples in RGB modality on NV-Gesture test dataset. (a)  Neural activation of I3D. (b)  Neural activation of RAAR3DNet.}
\label{fig:neur_act}
\end{figure*} 

\subsection{\textit{Feature Visualization of the RAAR3DNet}}
The feature maps (before Max-pool $3\times3\times3 $ in RAAR3DNet) are visualized in Fig.\ref{fig:neur_act}. Comparing with the original I3D network, it can be seen that the proposed RAAR3DNet are able to capture the movement information from hands and arms,  and gesture-irrelevant regions in the feature maps is decreased significantly due to the benefit from the searching and reconstructing the internal structure of a network architecture. 

% \begin,{figure*}[ht]
% \centering
% \subfigure[]{\includegraphics[width=0.45\linewidth]{img/I3D_feature1.png}}
% \subfigure[]{\includegraphics[width=0.45\linewidth]{img/hsdatt_I3D_feature1.png}}

% \subfigure[]{\includegraphics[width=0.45\linewidth]{img/I3D_feature2.png}}
% \subfigure[]{\includegraphics[width=0.45\linewidth]{img/hsdatt_I3D_feature2.png}}

% \subfigure[]{\includegraphics[width=0.45\linewidth]{img/Aut_features1.png}}
% \subfigure[]{\includegraphics[width=0.45\linewidth]{img/HSDAtt_Aut_features1.png}}

% \subfigure[]{\includegraphics[width=0.45\linewidth]{img/Aut_features2.png}}
% \subfigure[]{\includegraphics[width=0.45\linewidth]{img/HSDAtt_Aut_features2.png}}

% \subfigure[Heatmap]{\includegraphics[width=0.325\linewidth]{img/heatmap1.jpg}}
% \caption{Visualization results of the feature maps of the first 10 samples in the NV-Gesture test dataset. (a) and (b) are the low-level feature maps of the I3D network without DSA module and with DSA module. (c) and (d) are the corresponding high-level feature maps. (e) and (f) are the low-level feature maps of the NI3D network without DSA module and with DSA module. (g) and (h) are the corresponding high-level feature maps}
% \label{fig:neur_act}
% \end{figure*}

\bibliographystyle{aaai21}
\bibliography{gesture21}